\documentclass{article}

\usepackage[utf8]{inputenc}
\usepackage{url}
\usepackage[numbers,sort]{natbib}
\usepackage{fixltx2e}
\usepackage{booktabs}
\usepackage{graphics, graphicx, subfigure}
\usepackage{amsmath,amssymb}
\usepackage{xspace}
\usepackage{tikz}
\usetikzlibrary{arrows, decorations.markings,snakes}
\tikzstyle{vecArrow} = [thick, decoration={markings,mark=at position
   1 with {\arrow[semithick]{open triangle 60}}},
   double distance=1.4pt, shorten >= 5.5pt,
   preaction = {decorate},
   postaction = {draw,line width=1.4pt, white,shorten >= 4.5pt}]
\tikzstyle{vecArrow*} = [thick, decoration={markings,mark=at position
   1 with {\arrow[semithick]{open triangle 60}}},
   double distance=1.4pt, shorten >= 5.5pt,
   preaction = {decorate},
   postaction = {draw,line width=1.4pt, white,shorten >= 4.5pt}, 
   to path={-- node[inner sep=5pt,at end,sloped] {\huge${}^*$} (\tikztotarget) \tikztonodes}]
\tikzstyle{innerWhite} = [semithick, white,line width=1.4pt, shorten >= 4.5pt]
\usetikzlibrary{shapes.misc}
\tikzset{cross/.style={cross out, draw=black, minimum size=2*(#1-\pgflinewidth), inner sep=0pt, outer sep=0pt},
cross/.default={1pt}}

\usepackage{booktabs}

\newcommand{\x}{x}
\newcommand{\y}{y}
\newcommand{\X}{X}
\newcommand{\Y}{Y}
\newcommand{\e}{\varphi}
\newcommand{\p}{p}
\newcommand{\eparam}{\vartheta}
\newcommand{\defas}{\overset{\mathrm{def}}{=}}  %

\newcommand{\framework}{MONAS\xspace}

\usepackage{parskip}

\DeclareMathOperator*{\expectation}{\mathbb{E}}

\usepackage[verbose=true,letterpaper]{geometry}
\AtBeginDocument{
  \newgeometry{
    textheight=9in,
    textwidth=6.5in,
    top=1in,
    headheight=12pt,
    headsep=25pt,
    footskip=30pt
  }
}

\def\<{\begin{equation}}
\def\>{\end{equation}}  
\usepackage[toc,page]{appendix}
\usepackage[mathcal]{eucal}

\title{Automatic design of novel potential 3CL\textsuperscript{pro} and PL\textsuperscript{pro} inhibitors}
\author{Timothy Atkinson \and Saeed Saremi \and Faustino Gomez \and Jonathan Masci}
\date{%
    NNAISENSE S.A. \\
    June 2020 \\
}
\begin{document}

\maketitle

\begin{abstract}
    With the goal of designing novel inhibitors for SARS-CoV-1 and SARS-CoV-2, we propose the general molecule optimization framework, \textbf{Mo}lecular \textbf{N}eural \textbf{A}ssay \textbf{S}earch (\framework), consisting of three components: a property predictor which identifies molecules with specific desirable properties, an energy model which approximates the statistical similarity of a given molecule to known training molecules, and a molecule search method. In this work, these components are instantiated with graph neural networks (GNNs), Deep Energy Estimator Networks (DEEN) and Monte Carlo tree search (MCTS), respectively. This implementation is used to identify 120K molecules (out of 40-million explored) which the GNN determined to be likely SARS-CoV-1 inhibitors, and, at the same time, are statistically close to the dataset used to train the GNN.
\end{abstract}


\section{Introduction}

Over the past year, 
the search for molecules which may inhibit key receptor sites of Severe Acute Respiratory Syndrome Coronavirus-2 (SARS-CoV-2) has emerged as a central research objective within the scientific community~\cite{JEDI}. The already widespread use of Deep Learning (DL) techniques as predictors in molecular chemistry~\cite{gawehn2016deep,goh2017deep} has facilitated their rapid redeployment to this task e.g. \cite{butt2020deep,beck2020predicting,zhavoronkov2020potential} (see~\cite{latif2020leveraging} for a comprehensive review). There are a number of structural similarities between both the main 3CL\textsuperscript{pro} protease and the PL\textsuperscript{pro} protease of SARS-Cov-2, both used in reproduction, and those of Severe Acute Respiratory Syndrome Coronavirus-1 (SARS-CoV-1)~\cite{macchiagodena2020inhibition}. Based on these similarities, it has been suggested that the lack of available large-scale data for SARS-Cov-2 inhibition can be overcome using existing datasets of SARS-CoV-1 inhibitors~\cite{hofmarcher2020large}.

\begin{figure}
    \centering
    \input{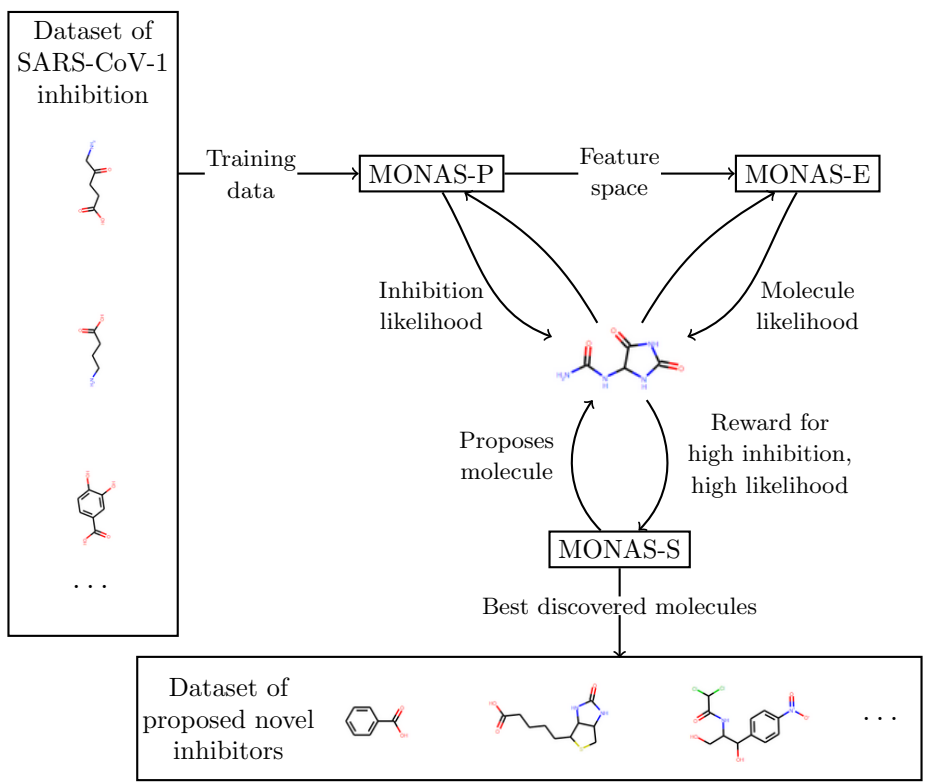}
    \caption{An overview of the \framework approach, applied to SARS-CoV-1 inhibition.}
    \label{fig:Approach}
\end{figure}

This paper proposes \textbf{Mo}lecular \textbf{N}eural \textbf{A}ssay \textbf{S}earch (\framework), a general data-driven automatic molecule design framework consisting of three core components (Figure \ref{fig:Approach}): 

\begin{itemize}
    \item \textit{\framework-P}: Property predictor.
    A predictive model of molecular activity which is used to identify new molecules with desirable properties. 
    \item \textit{\framework-E}: Energy Model. A general probabilistic model that learns the distribution of molecules in an unsupervised manner. This {\em energy} model assigns a scalar (energy) to any molecule and effectively provides a measure of statistical similarity between generated molecules and molecules in the dataset. 
    \item \textit{\framework-S}: Search. A search procedure to explore over the space of possible molecules that is guided by the inhibition predictor and the energy model. This component discovers new molecules to which the other two models assign both a high likelihood of inhibition with a high similarity to known molecules. 
\end{itemize}

In the particular instantiation of the framework used in this work, \framework-P is a Graph Neural 
Network (GNN;~\cite{Scarselli2009TheGN,battaglia2018relational,bronstein2017gdlsurvey,hamilton2017reprlearninggraphs}), the \framework-E is a Deep Energy Estimator Network~\cite{saremi2018deep, saremi2019neural}, and Monte Carlo Tree Search (MCTS;~\cite{browne2012survey}) is used as \framework-S to search over derivations of molecules in a Backus-Naur form grammar~\cite{kraev2018grammars}. However, we stress that this is only one possible implementation of this general framework. For example, \framework-P could instead use LSTM to predict behaviour from SMILES strings~\cite{mayr2018large}, DEEN could be replaced with a variational autoencoder~\cite{kingma2013auto} for \framework-E, and \framework-S could be performed by a genetic algorithm~\cite{jensen2019graph} or deep reinforcement learning~\cite{staahl2019deep}. Further, while in this work we focus particularly on inhibition of SARS-CoV-1 and SARS-CoV-2, in general this framework may be used for any data-driven molecular optimization task.

Existing approaches to the identification of SARS-CoV-2 inhibitors and reactants generally use DL to predict inhibition in order to screen sets of known inhibitors~\cite{beck2020predicting,hofmarcher2020large,ton2020rapid,zhang2020deep,fischer2020potential}. In contrast, we propose a \textit{de novo} approach to identifying and designing novel inhibitors for SARS-CoV-2.
There are two main motivations for doing this:
\begin{enumerate}
    \item The combinatorial space of small drug-like molecules is huge (approx. $10^{33}$ \cite{polishchuk2013estimation}) so that it is unlikely that even datasets with billions of molecules will contain the best SARS-CoV-2 inhibitors. Therefore, directed search through the full space may yield higher-quality results.
    \item It allows the system to discover potential inhibitors which are previously unknown molecules and therefore not patented. If any discovered molecule is eventually tested and found viable for treatment, this may substantially improve the general availability of treatment. 
\end{enumerate}

Unlike other de novo approaches which utilize docking simulations~\cite{cofala2020evolutionary} or structural information~\cite{tang2020ai}, \framework maximizes the output of a neural model trained to predict SARS-CoV-1 inhibition. The use of a predictive network greatly reduces the cost of sampling so that millions of molecules can be explored efficiently.

The rest of this work is organized as follows. In Section~\ref{sec:data}, the four datasets used to train a GNN-based Inhibition Predictor are described. The core concepts of GNNs and the specific architecture used  for \framework-P are described in Section~\ref{sec:GNN}. A brief overview of neural empirical Bayes which is used to learn an energy function on the feature space of the GNN (i.e.\ \framework-E) is provided in Section~\ref{sec:deen}. The directed search \framework-S over molecules, guided by these neural models, is described in Section~\ref{sec:mcts}. The performance of the neural models and the results of inhibitor design experiments are provided in Section~\ref{sec:experiments}. Finally, our findings are summarized in Section~\ref{sec:conclude}.

\section{Data \label{sec:data}}

As in~\citet{hofmarcher2020large}, four publicly available assays, taken from PubChem~\cite{kim2019pubchem}, were used. Each assay provides a set of molecules (represented by their SMILES strings~\cite{weininger1988smiles}) and the results of tests to determine whether or not each molecule inhibits a given protease for SARS-CoV-2. Each assay can be viewed as a mapping 
from its associated set of molecules $M_A$ to the binary vector $\{0, 1\}$ where $0$ indicates no inhibition and $1$ indicates inhibition. A summary of the assays used is provided in Table~\ref{tab:assay_summary}. 

\begin{table}[]
    \centering
    \begin{tabular}{l c rr}
        \toprule
        Assay ID & Target protease & \#Inactive molecules & \#Active molecules\\
        \midrule
        1,706 & 3CL\textsuperscript{pro} & 290,321 & 405 \\
        1,879 & 3CL\textsuperscript{pro} & 244 & 136 \\
        485,353 & PL\textsuperscript{pro} & 322,433 & 602 \\
        652,038 & PL\textsuperscript{pro} & 735 & 198\\ \midrule
        Total & - & 331,480 & 1,095\\
        \bottomrule
    \end{tabular}
    \caption{{\bf Assays used throughout experiments}. All assays are taken from PubChem \cite{kim2019pubchem}. The calculation of the totals takes into account that the same molecule may appear in multiple assays. The total number of active molecules is the number of molecules which are active in at least one assay.}
    \label{tab:assay_summary}
\end{table}

To use these assays with a GNN, each molecule is then formatted as a graph according to the following:
\begin{enumerate}
    \item For each molecule, convert its SMILES representation to a molecular representation using RdKit \cite{rdkit}.
    \item For each molecular representation:
    \begin{enumerate}
        \item For each atom, create a node with the following features: mass, valence, the total number of Hydrogens, whether it is aromatic, and formal charge. 
        \item For each bond between two atoms, create an edge in each direction between its corresponding nodes and additionally, create a self-loop on each node.\footnote{This is done for technical reasons with the GNN: it ensures that each node is treated as a member of its own neighborhood.} Each edge is categorized as one of: single bond, double bond, triple bond, aromatic bond or whether it is one of the synthetically added self-loops. 
    \end{enumerate}
    \item Each graph is assigned four binary classifications: one for each of the four assays, maintaining missing values.
\end{enumerate}


\section{Graph neural networks \label{sec:GNN}}

Graph neural networks are a relatively recent form of deep learning architecture for reasoning about structured relational data. As molecules are naturally structured, with atoms as nodes and bonds as edges, there is a clear affinity between graph neural networks and predictive molecular chemistry that has led to a number of developments in recent years~\cite{gilmer2017neural,hy2018predicting,st2019message}. In this work, the predictive component \framework-P is instantiated as a GNN architecture.

\subsection{GNN layers}

While there are a variety of unique GNN designs (see~\cite{zhou2018graph}), a typical construction is a layer which takes a directed labelled multi-graph $G_K = (V_K, E_K)$ and computes new node and edge features based on the existing features and adjacency, yielding a new graph $G_{K+1} = (V_{K+1}, E_{K+1})$ with the same structure but new labels. Here, the set $V = \{v_i\}_{i=1:N_V}$ is the set of $N_V$ nodes where each $v_i$ is the $i$th node's features, and $E_K = (e_j, s_j, t_j)_{j=1:N_E}$ is the set of $N_E$ edges where each $e_j$ is the $j$th edge's features, $s_j \in V_K$ is the $j$th edge's source node and $t_j \in V_K$ is the $j$th edge's target node. The transformed $G_{K+1}$ can then be passed to another graph neural network layer, have its node features used to classify nodes~\cite{perozzi2014dw,kipf2017semi} or predict edges~\cite{wang2018dyngraph,kipf2018neural}, or features can be pooled and passed to some other neural network to perform classification
or regression on the entire input graph~\cite{zhang2019hierarchical,DuvenaudMAGHAA15}.

In the following it will be helpful to refer to the \textit{neighborhood} of a given node $v_i$, e.g.\ the set of edges which target it. This is defined by $
    \mathcal{N}_i = \{j \mid e_j \in E \text{ and } t_j = v_i\}.$ In this work a simple model of graph neural network layers of the form in~\citet{battaglia2018relational} is used. To compute $G_{K+1}$ from $G_K$, first compute $E_{K+1}$, given by $E_{K+1} = \{e_j', s_j, t_j\}_{j=1:N_E},$ where $e_j'$ is a function of each edge's features, source node features and target node features:
    $e_j' = \Phi_E(e_j, v_{s_j}, v_{t_j}),$ where $\Phi_E$ is a multi-layer perceptron (MLP). Then the mean of the new edge features of the neighborhood of each node $v_i$ is computed as,
\begin{equation}
    \overline{\mathcal{N}}_i = \frac{1}{| \mathcal{N}_i |} \sum_{j \in \mathcal{N}_i} e_j'.
\end{equation}
The new node features $V_{K+1}$ are then computed as a function of each node's features and its mean aggregated neighborhood,
\begin{equation}
    V_{K+1} = \{v_i'\}_{i=1:N_V},
\end{equation}
\begin{equation}
    v_i' = \Phi_V(v_i, \overline{\mathcal{N}}_i),
\end{equation}
giving the updated graph $G_{K+1}$. As before, $\Phi_V$ is a multi-layer perceptron. An important property of the above construction is that, because each update is in the context of a node or edge's adjacency, and the permutation invariant mean aggregation of the neighborhood is used, the entire layer is invariant to permutations of the node and edge sets. Therefore, two isomorphic graphs will be updated in the same manner irrespective of the order in which the nodes (and edges) are indexed.

When classifying graphs, it is often helpful to `coarsen' the graph by merging nodes~\cite[Section V.C]{zhou2018graph} to reduce the number of parameters and avoid over-fitting. In this work the edge contraction pooling operator EdgePool \cite{diehl2019edge} is used to achieve this effect, which is extended to support edge features. EdgePool allows the graph coarsening to be learned through  parameters $W$ and $b$, such that a raw score can be assigned to each edge $e_{j}$ from source node $s_j$ to target node $t_j$ as,
\begin{equation}
    r(e_j) = W \cdot (s_j \oplus t_j \oplus e_j) + b,
\end{equation}
where $\oplus$ denotes concatenation. This is transformed into a score value $s(e_j)$ by taking the softmax of the neighborhood of source node $s_j$:
\begin{equation}
    s(e_j) = \text{softmax}_{\mathcal{N}_{s_j}} r(e_j).
\end{equation}

Edges are then iteratively contracted according to their score. 
That is,
starting at the edge with the highest score and continuing in descending order, contract an edge if and only if its source and target nodes have not been involved in any other edge contraction. When an edge $e_j$ is contracted, it is removed, its source node $s_j$ and its target node $t_j$ are merged, replacing all 3 items with a single node with the average of the node features multiplied by the 
score edge's associated score:
\begin{equation}
    v_j' = s(e_j) \frac{s_j + t_j}{2}. 
\end{equation}
Any incoming or outgoing edges of either $s_j$ or $t_j$ now are instead connected to $v_j'$. Furthermore, any edges which have become parallel as a result of this merge have their features merged, again by averaging. 
The result of these steps is a process whereby the graph is coarsened in a learned fashion while connectivity is preserved. In contrast, in alternative learnable pooling approaches such as Top-$K$ pooling~\cite{cangea2018towards} and Self-attention Graph pooling~\cite{lee2019self}, the fact that nodes are dropped, rather than merged, means that connectivity is lost. 

\subsection{\framework-P architecture}

\begin{figure}
    \centering
    \begin{tikzpicture}
\begin{scope}[every node/.style={rectangle,thick,draw}]
    \node[align=center,draw=none] (X) at (3, -0.5) {$G$};
    \coordinate (GEC1con) at (3, -1.25);
    \node[align=center] (GEC1) at (3, -2) {GEC\textsubscript{1}};
    \coordinate (GEC2con) at (3, -3);
    \node[align=center] (GEC2) at (3, -4) {GEC\textsubscript{2}};
    \coordinate (GEC3con) at (3, -5);
    \node[align=center] (GEC3) at (3, -6) {GEC\textsubscript{3}};
    \coordinate (GEC4con) at (3, -7);
    
    
    \coordinate (GEC1L) at (0, -3);
    
    \draw (0,-5) circle (6pt);
    \coordinate (GEC2Lcon) at (0, -4);
    \node[cross=5pt,rotate=45,black] (GEC2L) at (0, -5) {};
    
    \draw (0,-7) circle (6pt);
    \coordinate (GEC3Lcon) at (0, -6);
    \node[cross=5pt,rotate=45,black] (GEC3L) at (0, -7) {};
    
    \coordinate (GEC4Lcon) at (0, -8);
    \coordinate (headcon) at (0,-9.5);
    
    \coordinate (deenarr) at (0,-10.75);
    \node[align=center] (deen) at (0,-12) {DEEN};

    \coordinate (csrc1) at (3.6, -2);
    \coordinate (ctrg1) at (8.5, -1.9);
    \coordinate (ctrg2) at (5.7, -5.5);
    \path [dashed] (csrc1) edge (ctrg1);
    \path [dashed] (csrc1) edge (ctrg2);
    \draw[fill=white] (8.4, -4.75) circle (82pt);
    
    \coordinate (poolarr) at (0.75, -9.5);
    
    \coordinate (poolsplit) at (1.5, -9.5);
    \coordinate (poolsplittop) at (1.5, -8);
    \coordinate (poolsplittop2) at (1.5, -9);
    \coordinate (poolsplitbot2) at (1.5, -10);
    \coordinate (poolsplitbot) at (1.5, -11);
    
    \node[align=center,minimum width = 4cm] (MLP1) at (4.5, -8) {MLP for Assay 1,706};
    \node[align=center,minimum width = 4cm] (MLP2) at (4.5, -9) {MLP for Assay 1,879};
    \node[align=center,minimum width = 4cm] (MLP3) at (4.5, -10) {MLP for Assay 485,353};
    \node[align=center,minimum width = 4cm] (MLP4) at (4.5, -11) {MLP for Assay 652,038};
    
    \coordinate (poolmergetop) at (7.5, -8);
    \coordinate (poolmergetop2) at (7.5, -9);
    \coordinate (poolmergebot2) at (7.5, -10);
    \coordinate (poolmergebot) at (7.5, -11);

    \draw (7.5,-9.5) circle (6pt);
    
    \node[cross=5pt,rotate=45,black] (poolmerge) at (7.5, -9.5) {};
    \coordinate (presig) at (8.25, -9.5);
    \coordinate (postsig) at (9.75, -9.5);
    
    \node[align=center,draw=none] (MX) at (10.75, -9.5) {$f(G)$};
    \node[align=center,draw=none] (EX) at (10.75, -12) {$\e(X)$};
    
    \node[align=center,draw=none] (V) at (7, -3) {$V_k$};
    \coordinate (Vcon) at (7, -3.5);
    \node[align=center,draw=none] (E) at (10, -3) {$E_k$};
    \coordinate (Econ) at (10, -3.5);
    \node[align=center, minimum height=0.6cm] (EMLP) at (10, -4) {\small Edge MLP};
    \node[align=center, minimum height=0.6cm] (VMLP) at (7, -4) {\small Node MLP};
    \coordinate (VEcon) at (8.5, -4);
    \coordinate (VTopKcon) at (7, -4.75);
    \coordinate (ETopKcon) at (10, -4.75);
    \coordinate (ETopKcon2) at (10, -5.5);
    \node[align=center, minimum height=0.6cm] (TopK) at (7, -5.5) {\small Edge Pooling};
    
    \node[align=center,draw=none] (Vpr) at (7, -6.5) {$V_{k+1}$};
    \coordinate (Vprcon) at (7, -6);
    \node[align=center,draw=none] (Epr) at (10, -6.5) {$E_{k+1}$};
    \coordinate (Eprcon) at (10, -6);
    
\end{scope}

\begin{scope}[every node/.style={fill=white,circle},
              every edge/.style={draw=black,thick}]
            \path [->] (X) edge (GEC1con);
            \path [] (GEC1con) edge (GEC1);
            \path [->] (GEC1) edge (GEC2con);
            \path [] (GEC2con) edge (GEC2);
            \path [->] (GEC2) edge (GEC3con);
            \path [] (GEC3con) edge (GEC3);
            \path [] (GEC3) edge (GEC4con);
            
            \path [] (GEC2con) edge node {\small Mean/Max} (GEC1L);
            \path [] (GEC3con) edge node {\small Mean/Max} (GEC2L);
            \path [] (GEC4con) edge node {\small Mean/Max} (GEC3L);
              
            \path [->] (GEC1L) edge (GEC2Lcon);
            \path [] (GEC2Lcon) edge (GEC2L);
              
            \path [->] (GEC2L) edge (GEC3Lcon);
            \path [] (GEC3Lcon) edge (GEC3L);
            
            \path [->] (GEC3L) edge (GEC4Lcon);
            \path [] (GEC4Lcon) edge (headcon);
            
            \path [->] (headcon) edge (poolarr);
            \path [] (poolarr) edge (poolsplit);
            \path [] (poolsplittop) edge (poolsplitbot);
            
            \path [] (poolsplittop) edge (MLP1);
            \path [] (poolsplittop2) edge (MLP2);
            \path [] (poolsplitbot2) edge (MLP3);
            \path [] (poolsplitbot) edge (MLP4);
            
            \path [] (MLP1) edge (poolmergetop);
            \path [] (MLP2) edge (poolmergetop2);
            \path [] (MLP3) edge (poolmergebot2);
            \path [] (MLP4) edge (poolmergebot);
            
            \path [] (poolmergetop) edge (poolmerge);
            \path [] (poolmergebot) edge (poolmerge);
            
            \path [->] (poolmerge) edge (presig);
            \path [] (presig) edge node {\small Sigmoid} (postsig);
            \path [->] (postsig) edge (MX);
            
            \path [->] (headcon) edge node {$\X$} (deenarr);
            \path [] (deenarr) edge (deen);
            \path [->] (deen) edge (EX);
            
            
            \path [->] (V) edge (Vcon);
            \path [] (Vcon) edge (VMLP);
            \path [->] (E) edge (Econ);
            \path [] (Econ) edge (EMLP);
            \path [->] (EMLP) edge (VEcon);
            \path [] (VEcon) edge (VMLP);
            
            \path [->] (VMLP) edge (VTopKcon);
            \path [] (VTopKcon) edge (TopK);
            
            \path [->] (EMLP) edge (ETopKcon);
            \path [] (ETopKcon) edge (ETopKcon2);
            \path [] (ETopKcon2) edge (TopK);
            
            \path [] (TopK) edge (Vprcon);
            \path [->] (Vprcon) edge (Vpr);
            \path [] (Vprcon) edge (Eprcon);
            \path [->] (Eprcon) edge (Epr);
\end{scope}
\end{tikzpicture}
    \caption{\textbf{The multi-task graph neural network used to predict SARS-CoV-1 protease inhibition activity for molecules.} $G$ is an input graph. Each GEC block is decomposed into a GNN layer followed by edge contraction pooling with outputs passed to the next block. The output of each block is also passed through global mean and max pooling, with the results of all pooling concatenated ($\bigoplus$) together to form a vector representation of the input graph. This vector is passed through an MLP for each target assay, yielding a binary prediction for each task $f(G)$. The latent representation $\X$ is then passed to the Deep Energy Estimator Network (DEEN) to generate an energy $\varphi(X)$, see Section \ref{sec:deen}. }
    \label{fig:gnn}
\end{figure}
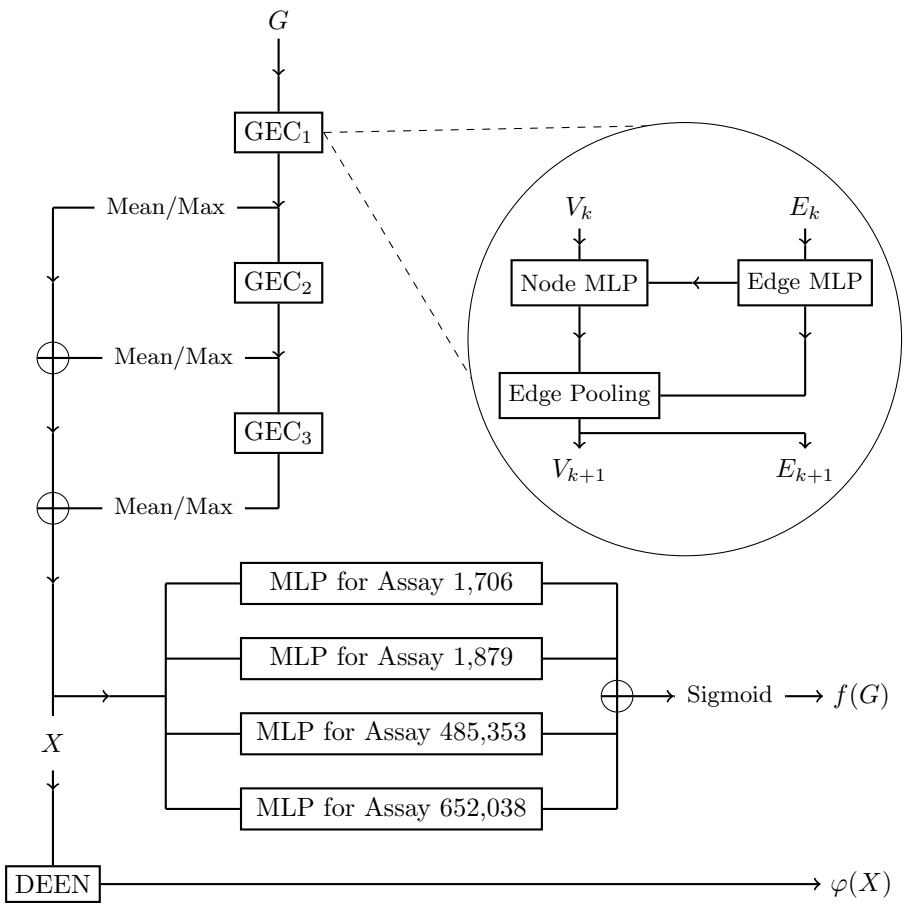

We treat prediction of inhibition as multi-task and create a single network which predicts inhibition for all four assays. 
The architecture is shown in Figure~\ref{fig:gnn}. Initially, the input graph $G$ is passed through three `GNN Layer - Edge Contraction' (GEC) blocks. Each GEC block is a graph neural network layer, consisting of node MLP $\Phi_V$ and edge MLP $\Phi_E$, followed by edge contraction pooling. In every GEC block, both $\Phi_V$ and $\Phi_E$ are 2-layer MLPs with $96$ neurons per layer, and batch normalization applied after each layer followed by ReLU activation. Each edge contraction pooling operator removes approximately half of the edges from the graph so that the graph, after the third GEC block, is expected to have $1/8$th of the edges of the input graph. 

After each GEC block, global mean and global max pooling are applied to the $96$-dimensional node features, yielding a $192$-dimensional representation of the input. The three representations from the GEC blocks are concatenated together, yielding the $574$-dimensional latent representation of the input, denoted $\X$.

For each assay, a `head' is used, which is a specialist MLP that predicts only inhibition for that assay. Each head is provided with the latent representation $\X$ and applies two $128$-neuron feed-forward layers, with dropout applied before each layer ($p=0.25$ and $p=0.5$, respectively), and batch normalization and ReLU activation after. Then, the $128$-dimensional vector is passed to a single logit neuron which gives a prediction for the given assay. The predictions are concatenated and passed through a softmax, yielding the prediction vector $M(G)$. 

\subsection{Inhibition loss \label{sec:loss}}

To handle the multi-task context while maintaining efficiency, the collective $331,480$ molecules are treated as a single dataset and are passed through the GNN in a mini-batch fashion. However, adjustments are made to a standard binary cross-entropy loss function to ensure that the network is not biased towards any given assay and that missing data is appropriately handled. For a given graph (molecule) $G$ with ground truth $K_A$ for assay $A$, the loss is computed for the $A$-th head of the model, $f_A(G)$, as

\begin{equation} \label{eq:lossA}
    \ell(K_A, f_A(G)) = \begin{cases}
        - \alpha_A \beta_A \log{(f_A(G))}, & K_A = 1\\
        -\alpha_A \log{(1 - f_A(G))}, & K_A = 0\\
        0, & K_A \text{ is missing,}\\
        \end{cases}
\end{equation}

where the loss for one single molecule/graph across all assays is given by 
\begin{equation}
    \ell(K, G) = \sum_A \ell(K_A, f_A(G)),
\end{equation}
     and the total loss $\ell(\theta)$ to be minimized, 
\begin{equation}
    \ell(\theta) = \expectation_{(K,G)} \ell(K, G),
\end{equation}
      where $\theta$ are the parameters of the GNN (omitted from Eq.~\ref{eq:lossA} for cleaner notation).

The factors $\alpha_A$ and $\beta_A$ are weights towards each task and positive instances of that task, respectively to handle the extreme imbalances of the data (see Table \ref{tab:assay_summary}). For assay $A$ with $I$ training inhibitors, $J$ training non-inhibitors and $N$ total training samples, $\alpha_A = N/(I + J),$ and, $\beta_A = (I + J)/I$. The role of $\alpha_A \geq 1$ is to ensure that each assay contributes equally to the total loss averaged across all $N$ training samples, whereas $\beta_A \geq 1$ ensures that positive and negative samples for each assay contribute to the total loss averaged across all $I + J$ training molecules for assay $A$.

\section{\framework-E: Deep Energy Estimator Network \label{sec:deen}}

Constructing a probabilistic model for the molecules and a statistical measure of how similar two molecules are based on dataset samples faces two main challenges: (i) In high dimensions,  classical techniques like kernel density estimation break down because the volume of space grows exponentially (i.e.\ curse of dimensionality) and nonparametric methods which are based on Euclidean similarity metric become inadequate. (ii) One typically resorts to parameterizing distributions in directed or undirected graphical models, but, in both cases, the inference over latent variables is in general intractable. At the root of the problem lies in estimating the \emph{normalizing constant} (a.k.a. the partition function) of probability distributions in high dimensions.


There is however a class of models, called \emph{energy models}, that are formulated around learning \emph{unnormalized} densities or energy functions. For MCTS, where only the likelihood of generated molecules  \emph{relative} to the likelihood of the molecules in the dataset is of importance (this will become clearer in the next section), the distribution needs only be known up to a constant. Therefore, learning energy functions is sufficient. In this section, we review a recent development~\cite{saremi2019neural} that formulates the problem of learning energy functions in terms of the empirical Bayes methodology~\cite{robbins1956empirical,miyasawa1961empirical}. The framework is referred to by "DEEN" due to its origin in Deep Energy Estimator Networks~\cite{saremi2018deep}. 

\begin{figure}[t]
    \centering
    \includegraphics[width=0.4\textwidth]{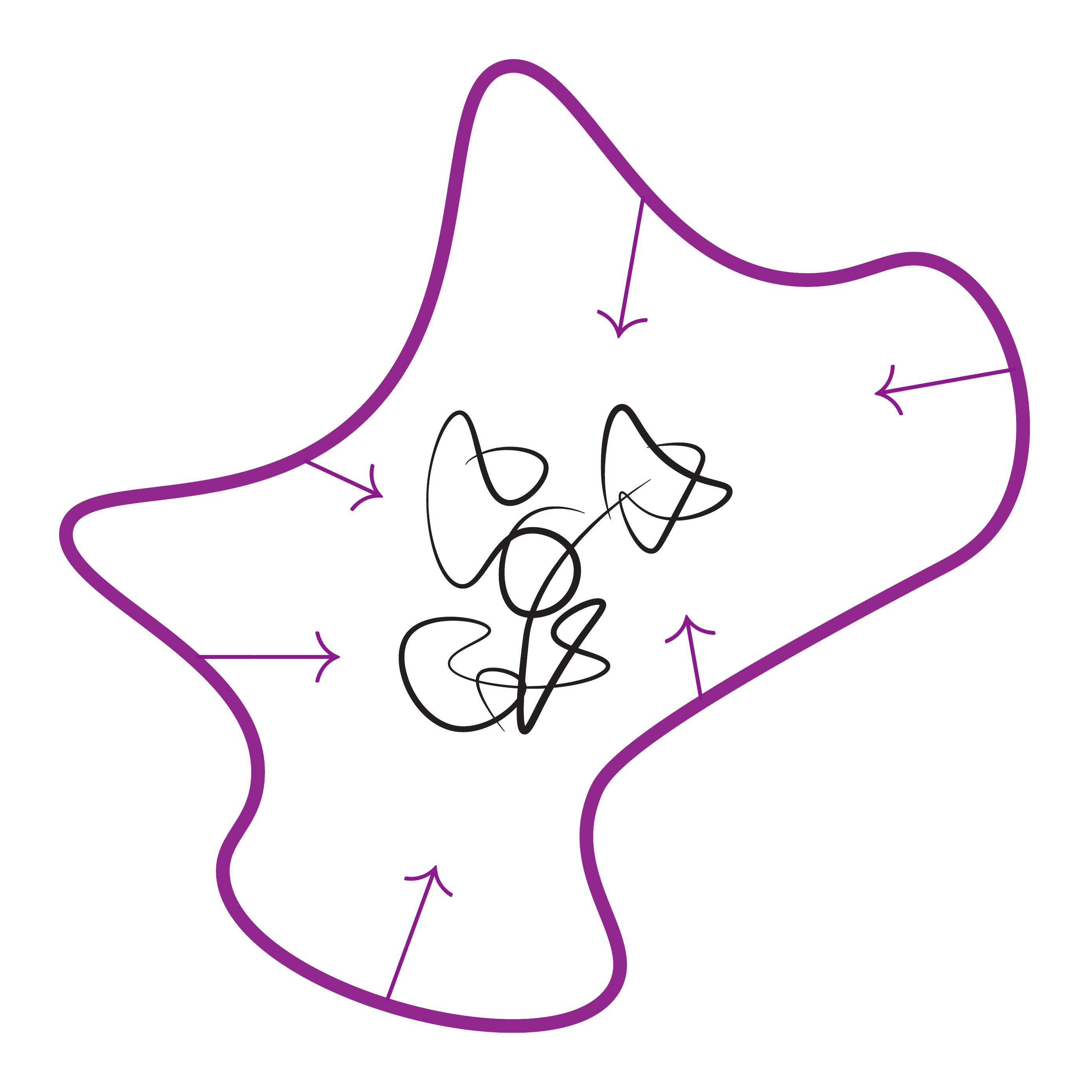}
    \caption{{\bf Illustration of DEEN process}.  Samples from the random variable $X$ (in the feature space of molecules) are represented in black, noisy samples from $Y=X+N(0,\sigma^2 I_d)$ in violet, and the arrows represent $-\nabla \e(y)$, where $\e$ is the energy function
    implemented by a neural network. DEEN's learning objective is a denoising objective rooted in empirical Bayes: the idea is to shape the energy function such that the arrows point to the data manifold.}
    \label{fig:deen}
\end{figure}

In this work, the problem is further simplified by learning the energy function on the feature space $\mathcal{X}=\mathbb{R}^{d_X}$ of the GNN (here $d_X=4018$). Instead of learning a probabilistic model for the i.i.d. sequence $G_1,\dots,G_n$, the representation of the sequence in the feature space $X_1,\dots,X_n$ is studied instead (see Figure~\ref{fig:gnn}).  
This simplification is due to technical reasons, since the denoising methodology of empirical Bayes is formulated in the Euclidean space where specifically the isotropic Gaussian $N(0,\sigma^2 I_d)$ that defines the \emph{noise model} plays a central role. 
Geometrically, the model is designed such that the negative gradient of the energy function evaluated on the noisy data is directed towards the clean data (see Fig.~\ref{fig:deen}). In other words, learning can be viewed as ``shaping'' the energy function $\e$ such that $-\nabla \e$ (known as the score function~\cite{hyvarinen2005estimation}) points toward the data manifold.

Some technical aspects of DEEN are reviewed next. Consider the Gaussian noise model and corrupt the i.i.d. samples $X_1,\dots,X_n$ as follows:
\begin{equation} 
\label{eq:Yij} \Y_{ij} = \X_i + \varepsilon_j,\text{ where } \varepsilon_j \sim N(0,\sigma^2 I_d). 
\end{equation}  
An important result in empirical Bayes is that the Bayes estimator of $\X$, given a noisy measurement $\Y=\y$, is given by $\hat{\x}(\y) = \y + \sigma^2 \nabla \log \p(\y)$, where $\nabla$ is the gradient with respect to the noisy data $\y$, and $\p(\y)$ is the probability density function of the random variable $Y=X+N(0,\sigma^2 I_d)$. The key step in DEEN is to parameterize the energy function of $Y$ with a neural network $\e_\eparam:\mathbb{R}^d \rightarrow \mathbb{R}$, where the Bayes estimator takes the parametric form: 
\begin{equation}
    \hat{\x}_{\eparam}(\y) = \y - \sigma^2 \nabla \e_\eparam(\y).
\end{equation}
Since the Bayes estimator is the least-squares estimator, the
learning objective follows immediately:
\begin{equation}
    \mathcal{L}(\eparam) = \mathbb{E}_{(\x,\y)} \Vert \x - \hat{\x}_{\eparam}(\y)  \Vert_2^2,
\end{equation}
where the expectation is over the empirical distribution over the samples $(\X_i,\Y_{ij})$ (see Eq.~\ref{eq:Yij}) and $\Vert \cdot \Vert_2$ is the Euclidean norm.  The main appeal of the algorithm is that it transforms learning the energy function into an optimization problem because it sidesteps posterior inference or MCMC approximations in latent variable models~\cite{wainwright2008graphical}.  At optimality, given a
neural network with sufficient capacity, and enough unlabelled samples ($n \gg 1$), finding a good approximation to the energy function: $\e(y,\eparam^*) \approx - \log p(y)$ is theoretically guaranteed (modulo a constant).

Summarizing, in terms of learning the distribution of graph-valued molecules, three simplifications are made:

\begin{enumerate}
    \item The problem is transformed into learning the distribution in the Euclidean space $\mathcal{X}$, defined by the GNN classifier. This simplification is indeed well-suited in this context, and is in line with the desire of having the molecules generated by MCTS be classified as inhibitors.
    \item The statistical model is un-normalized, i.e., the energy function is defined modulo an additive constant. This is allowed since only the \emph{difference} between energies appear in the reward function (introduced next), and the constant cancels out. 
    \item The algorithm is designed around learning the \emph{smoothed}
    density associated with noisy data with the hyperparameter $\sigma$. This relaxation is in fact key for regularizing the learning~\cite{saremi2019neural}, but not critical in the MCTS reward function as will become clear.
\end{enumerate}

\begin{figure}
    \input{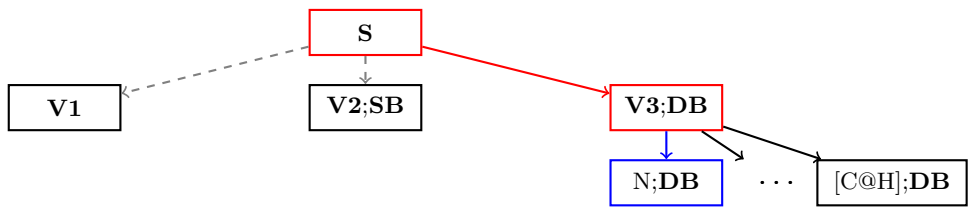}
    \caption{\textbf{An overview of the use of MCTS to generate molecules maximizing the reward function.}}
    \label{fig:mcts}
\end{figure}


\section{\framework-S: Designing candidate inhibitors with MCTS \label{sec:mcts}}

To automatically design new inhibitors, the conventional Upper Confidence Trees (UCT)
variant of MCTS~\cite{kocsis2006bandit} that searches over a BNF grammar of SMILES strings~\cite{kraev2018grammars}, is used\footnote{A definitive description of the algorithm can be found in~\citet{browne2012survey}}.  MCTS searches a tree of \textit{derivations}, i.e. sequences of steps which generate valid SMILES strings. In each iteration of MCTS, the following steps are taken (Figure~\ref{fig:mcts} depicts each of the four steps schematically):

\begin{enumerate}
    
    \item \textit{Selection}: starting from a root node with an associated initial non-terminal (`S'), successive child nodes are selected according to the UCB1 formula until a previously unvisited node is encountered. In Figure~\ref{fig:mcts} the root node is initially unexpanded, so it is selected. 
    \item \textit{Expansion}: the encountered node is expanded according to the possible productions from Leftmost parsing, yielding three children, `V1', `V2;SB' and `V3;DB', in the case of our example. One of the new children (`V2;SB') is selected at random. If the encountered node has no viable children, it represents a completed SMILES string, and is instead simply evaluated.
    
    \item \textit{Rollout}: a random derivation is generated for the selected child, such that the partially completed SMILES string at the selected node is completed at random yielding a valid SMILES string. In our case, the rollout is performed uniformly at
    random, except when a terminal symbol is available; in which case decisions are made uniformly at random over available terminal symbols. The  completed SMILES string is then converted into a molecular representation and evaluated using the reward function $w$.
    \item \textit{Backpropagation}: the evaluated reward $w_\beta(G)$ of the molecule $G$ is backpropagated through the tree. Every node that was visited has its average reward $\bar{w}$ and maximum reward $w^{\text{max}}$ updated. Additionally, the number of visits to each visited node $n$ is incremented. 
\end{enumerate}
We use a mild adjustment of the UCB1 (Upper Confidence Bound 1 applied to trees) formula~\cite{auer2002finite} to
determine the appropriate node/production to use in each step of the selection process. The modified UCB1 formula used is
\begin{equation}
    \frac{w_i^{\text{max}} + \bar{w}_i}{2} + c \sqrt{\frac{\ln{N_i}}{n_i}},
\end{equation}
where for node $i$, $w_i^{max}$ is the maximum reward observed, $\bar{w}_i$ is the average reward observed, $n_i$ is the total number of samples, $N_i$ is the total number of samples at the parent of node $i$, and $c$ is the exploration/exploitation parameter. The introduction of a maximum reward term $w_i^{max}$ is simply to help the system pursue particularly promising samples which are otherwise avoided due to a few poor-quality samples from that same node. Once all non-terminals have been cleared either by selection or rollout, a complete SMILES string is generated for evaluation. As stated, the rollout policy is random, but prioritizes terminals where possible in order to bias our sampling towards smaller molecules.

Every time MCTS generates a SMILES string, it is converted to a molecule and then a graph, as described in Section~\ref{sec:data}. By passing the sample molecule $G$ through the GNN, a prediction of inhibition $f_A(G)$ is obtained for assay $A$, and a latent representation $X$ of the molecule which is further passed through the DEEN model to yield an energy $\e(X)$. With $\phi_{\rm min}$ defined as the minimum energy for any known inhibitor in the test set, the reward returned to MCTS is then

\begin{equation}\label{eqn:reward}
    w_\beta(G) \defas f_A(G) \cdot \frac{2}{1+\exp(\beta \Delta \e(X))},\text{ where } \Delta \e = \e(X)-\phi_{\rm min},
\end{equation}

meaning that the sample molecule is heavily penalized for either a low inhibition prediction or high energy. The $\beta$ term is a hyperparameter controlling the smoothness of the energy component of the reward function, with a default value of 
\begin{equation}
    \beta_{0} =\frac{1}{ \phi_{\rm max}-\phi_{\rm min}},
\end{equation} 
(maximum and minimum are computed over the test set). A simple but important property of the reward function is that it is invariant under
$\varphi \rightarrow \varphi + C$. The reward function must be invariant to this transformation since the energy function itself is defined modulo an additive constant. 


\section{Experiments \label{sec:experiments}}

The automatic molecule design framework comprises three steps:

\begin{enumerate}
    \item The ensemble of GNNs is trained to predict inhibition for the four assays (\S \ref{subsec:GNN_training}).
    \item DEEN is then trained on the feature space of molecules generated by the ensemble of GNNs to learn an energy model of the dataset (\S \ref{subsec:DEEN_training}).
    \item The GNN ensemble and the DEEN model are used in the reward function (Eq. \ref{eqn:reward}) to guide MCTS, generating potential novel inhibitors (\S \ref{subsec:MCTS_running}). 
\end{enumerate}

\subsection{Training the GNN Ensemble \label{subsec:GNN_training}}

\begin{figure}
    \centering
    \begin{subfigure}[Validation set]
    {\includegraphics[width=0.8\textwidth]{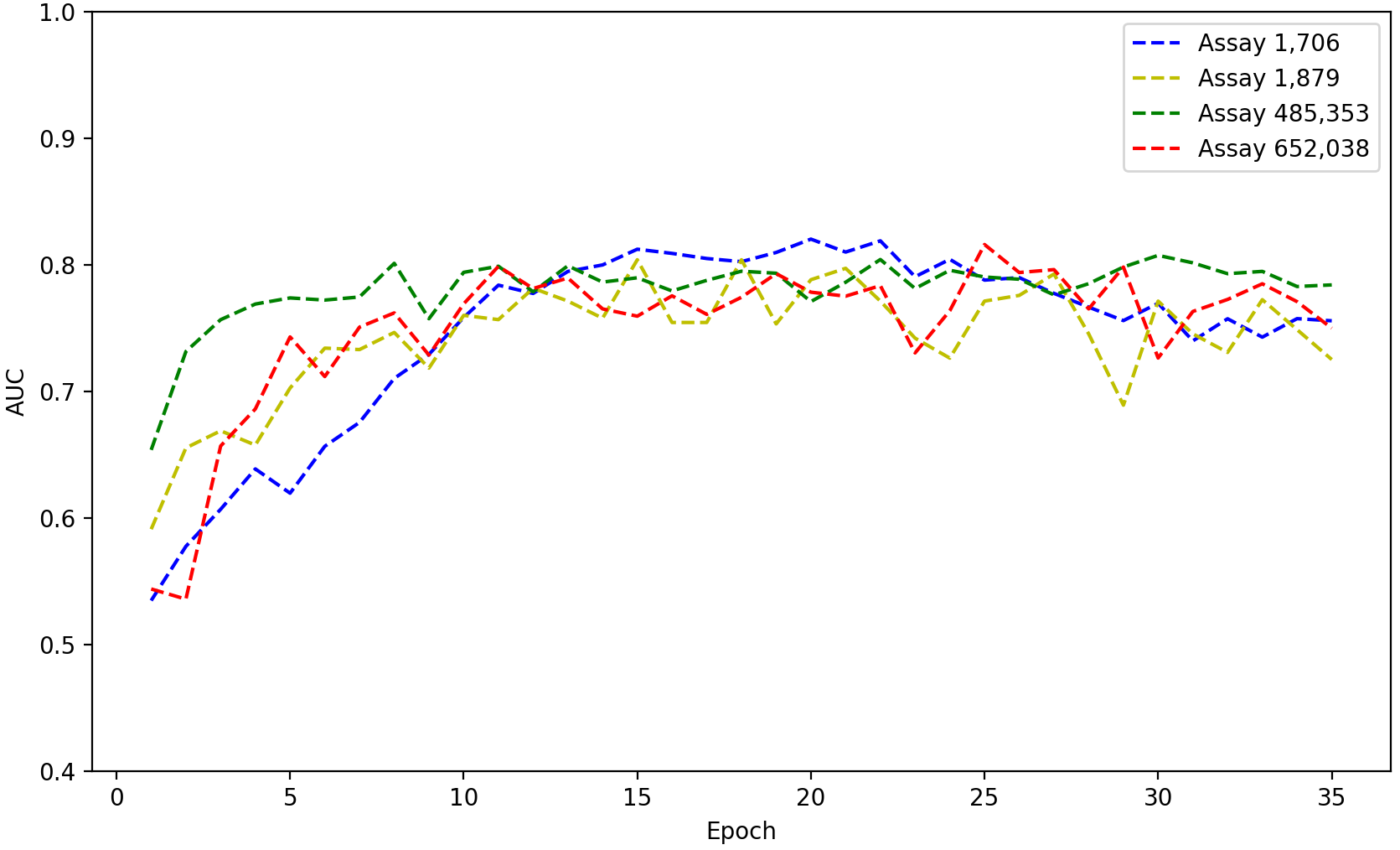}}
    \end{subfigure}\vspace*{2mm}
    \begin{subfigure}[Test set]
    {\includegraphics[width=0.8\textwidth]{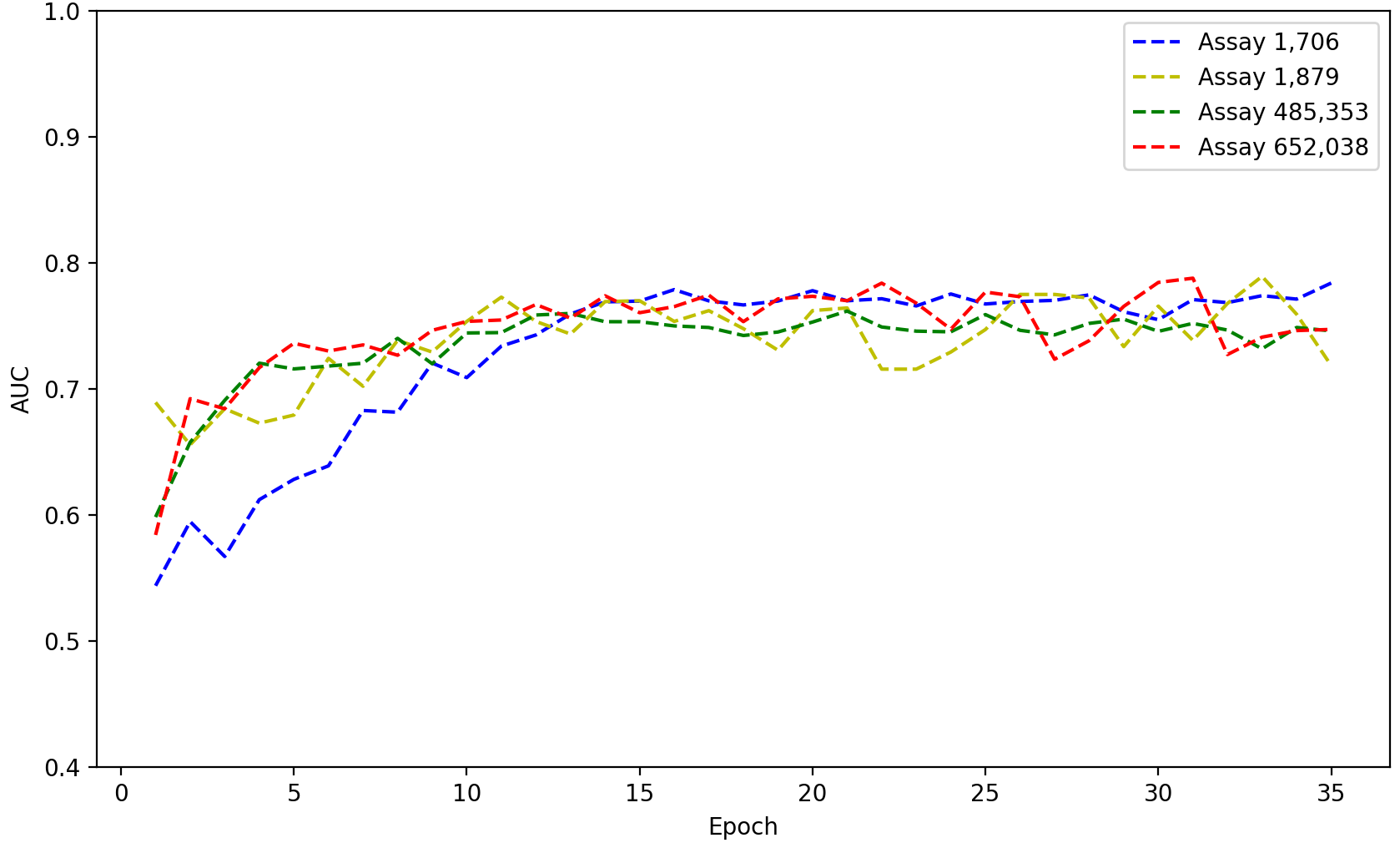}}
    \end{subfigure}
    \caption{\textbf{Validation and testing ROC AUC scores of model $0$} (see Table \ref{tab:roc_auc}), measured during training and reported for each assay. The validation loss is lowest at epoch $15$. }
    \label{fig:roc_auc_scores}
\end{figure}


To further boost the performance, given the relatively small number of positive samples, a simple Bootstrap-Aggregating ensemble of models~\cite{breiman1996bagging} was used in which the outputs of the ensemble members are averaged to generate predictions. To build the ensemble, the training data is split into five non-overlapping folds of equal size. Each model is trained on a different set of four folds in a standard cross-fold manner, yielding five models which have been trained under different conditions. For each model, the unused fold serves as a validation set. 
The parameters of each at the epoch at which the validation loss is lowest is used to construct the final ensemble. 

The data was first split into two disjoint subsets: 
$80\%$ of the data training, $20\%$ for testing. The training data is then split into five folds, and a network is trained on each subset of four folds, yielding five models. 
The various subsets of data are stratified such that each contains approximately the same number of positive and negative samples for each task.
Each model was trained using the ADAM optimizer~\cite{kingma2014adam}, with an initial learning rate of $2\times10^{-5}$, batch size of $128$ and weight decay of $1\times10^{-4}$. Each model was trained until the validation loss did not improved for $20$ epochs. Here, the validation loss is computed on the unused fold for each model.

A typical training cycle is shown in Figure~\ref{fig:roc_auc_scores}. The lowest validation loss for this model was observed at epoch $15$.  A full listing of ROC AUC scores of the models is given in Table~\ref{tab:roc_auc}, alongside those of the ensemble computed by averaging the outputs of the member models. The ensemble achieves higher scores than the average score of its individual members across all assays. In general, the ensemble has a $75-80\%$ likelihood of ranking an inhibitor higher than a non-inhibitor across all tasks. Although there is room for further improvement, meaningful classification performance is clearly 
demonstrated.

To validate the \framework-P architecture, it was compared empirically with
five other ensembles, each using a different type of GNN found in 
molecular chemistry literature: 
Message Passing Neural Networks (MPNNs;~\cite{gilmer2017neural}), Directed Message Passing Neural Networks 
(D-MPNN;~\cite{yang2019analyzing}), Attentive FingerPrints (Attentive FP;~\cite{xiong2019pushing}), and the full Graph Networks architecture of~\citet{battaglia2018relational} with global conditioning. The implementation details for each method are provided in the Appendix. 

The results of these comparisons are shown in Table~\ref{tab:gnn_compare}.
all for the same reserved subset of test data. 
\framework-P out-performs all other approaches studied, except on assay $652,038$ where it is comparable to Attentive FP and Graph Networks and is marginally outperformed by D-MPNN. Furthermore, \framework-P provides the highest ROC AUC score averaged across all four assays, confirming that it is well-suited to the problem. 
Note that the intent of these experiments is not to conduct an exhaustive empirical comparisons on this task, but is instead to provide context and verify that \framework-P is both appropriate and comparable to other state-of-the-art techniques.

\begin{table}[t!]
    \centering
    \begin{tabular}{l|c|rrrr}
        \toprule
        & Termination & \multicolumn{4}{c}{Assay ROC AUC Score} \\\cline{3-6}
        Model & Epoch & 1,708 & 1,879 & 485,353 & 652,038\\
        \midrule
        0 & 15 & 0.77 & 0.77 & 0.76 & 0.76 \\
        1 & 15 & 0.77 & \textbf{0.83} & 0.74 & 0.78 \\
        2 & 21 & \textbf{0.77} & 0.83 & \textbf{0.77} & \textbf{0.79} \\
        3 & 22 & 0.76 & 0.77 & 0.72 & 0.79 \\
        4 & 16 & 0.75 & 0.78 & 0.76 & 0.79 \\
        \midrule
        Average & 17.8 & 0.76 & 0.79 & 0.75 & 0.78 \\
        & & & & & \\
        Ensemble & - & \textbf{0.78} & \textbf{0.81} & \textbf{0.76} & \textbf{0.80} \\
        \bottomrule
    \end{tabular}
    \caption{\textbf{ROC AUC scores of the GNN ensemble computed on test data.} For each assay, the
    score of the best performing model is highlighted in
    \textbf{bold}. The average (mean) scores are provided in comparison that of the ensemble as a whole. For every assay, the ensemble performs better on the test data than the average of its individual members. In two assays (1,708 and 652,038), the ensemble performs better than \textit{any} of the individual GNNs.}
    \label{tab:roc_auc}
\end{table}
\begin{table}[t!]
    \begin{center}
    \begin{tabular}{l|rrrrr}
        \toprule
        & \multicolumn{5}{c}{Assay ROC AUC Score} \\\cline{2-6}
        Model & 1,708 & 1,879 & 485,353 & 652,038 & Average\\
        \midrule
        MPNN \cite{gilmer2017neural} & 0.75 & 0.78 & 0.74 & 0.78 & 0.76 \\
        D-MPNN \cite{yang2019analyzing} & 0.74 & 0.71 & 0.67 & \textbf{0.81} & 0.73 \\
        Attentive FP \cite{xiong2019pushing} & 0.76 & 0.75 & 0.73 & 0.80  & 0.76 \\
        Graph Networks \cite{battaglia2018relational} & 0.76 & 0.79 & 0.73 & 0.80 & 0.77 \\
        \textbf{\framework-P} & \textbf{0.78} & \textbf{0.81} & \textbf{0.76} & 0.80 & \textbf{0.79} \\
        \bottomrule
    \end{tabular}
    \end{center}
    \caption{\textbf{Inhibition prediction performance comparison}. For each assay the best performing model's ROC AUC score is highlighted in \textbf{bold}. Each result is reported for a bagging ensemble of five models each trained on different sub-sets of the training data.}
    \label{tab:gnn_compare}
\end{table}

\subsection{Training DEEN \label{subsec:DEEN_training}}

\begin{figure}
    \centering
    \includegraphics[width=0.8\textwidth]{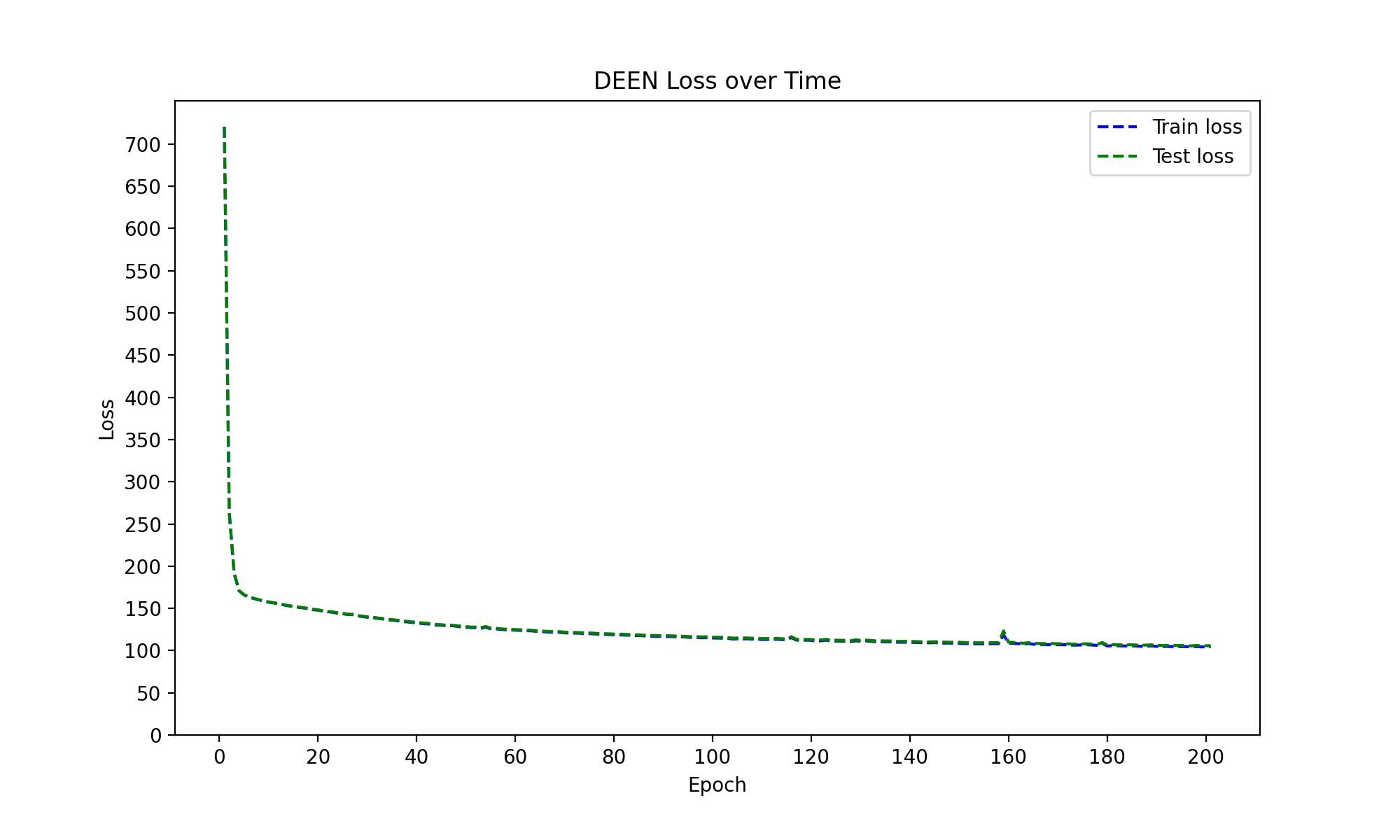}
    \caption{\textbf{DEEN model learning curve.} The loss for training and testing data remains similar throughout the training process.}
    \label{fig:deen_perf}
\end{figure}

DEEN is trained on the feature space of the training molecules with Gaussian noise with $\sigma=0.25$. The $574$-dimensional latent representations taken across all five members of the ensemble are concatenated to yield a single $2880$-dimensional latent representation (labeled $\X$ in Section~\ref{sec:deen}). 
An MLP was used with four layers of $3072, 2048$ and $1024$ neurons, respectively, followed by a single linear output neuron. Each layer, except for the first and last layer, uses a skip connection so that it takes, as input, the concatenation of the two previous layer outputs. All neurons use the smoothed ReLU activation function 
$u/(1+\exp(-u))$
which is known by two different names: \emph{SiLU}~\cite{elfwing2017sigmoid} and \emph{Swish}~\cite{ramachandran2017swish}. The choice of a \emph{smooth} activation function is important here\footnote{With ReLU, the optimizer saturates at significantly higher loss (compared to SiLU/Swish).} due to the double backpropagation (one in $y$ to compute the loss, the other in $\vartheta$ to compute the gradient of the loss) for a single SGD update.  The ADAM optimizer is again used, with a learning rate of $10^{-5}$ and a batch-size of $128$. DEEN is trained for $200$ epochs. 

Figure~\ref{fig:deen_perf} shows the training and testing loss measured during the $200$ training epochs. The epoch with the lowest testing loss is the final epoch. 
Note that the test loss closely follows the training loss, suggesting that the testing data lies on the same manifold of latent representations as the training data.

\subsection{Discovering novel inhibitors \label{subsec:MCTS_running}}

For each assay, MCTS optimizes the reward function for one million samples. This process is repeated ten times, yielding a total search of ten million molecules per assay. Once MCTS has terminated, the top $30$K unique sampled molecules are identified for each assay, where uniqueness is determined by comparing canonical SMILES strings. 
Additionally, in order to bias search toward smaller molecules, any derivation branch which reaches a depth $\ge 30$ immediately prioritizes terminals to pursue termination of the derivation. The exploration constant $c$ is set to $\sqrt{2}$. 

\begin{figure}[h!] 
\begin{center}
\begin{subfigure}[Known Inhibitors]
 {\includegraphics[width=0.72\textwidth]{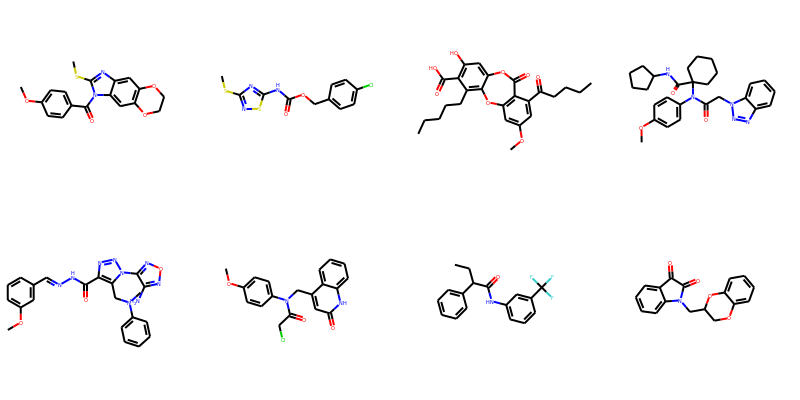}}
\end{subfigure}
\begin{subfigure}[Inhibitors discovered \emph{with} energy regularization ($\beta \neq 0$)]
 {\includegraphics[width=0.72\textwidth]{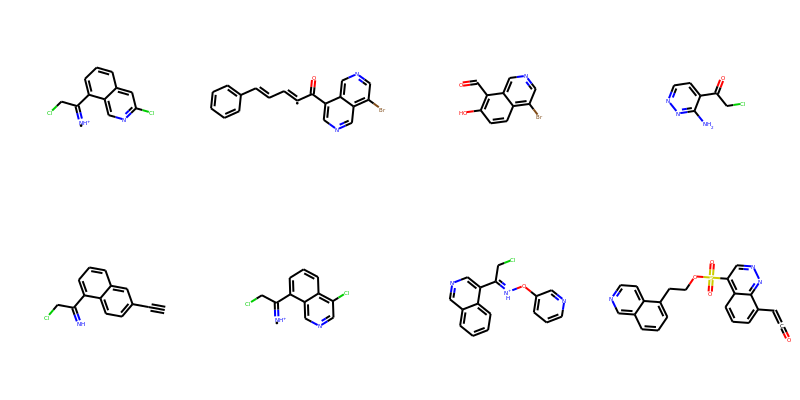}}
\end{subfigure}
\begin{subfigure}[Inhibitors discovered \emph{without} energy regularization ($\beta=0$)]
 {\includegraphics[width=0.72\textwidth]{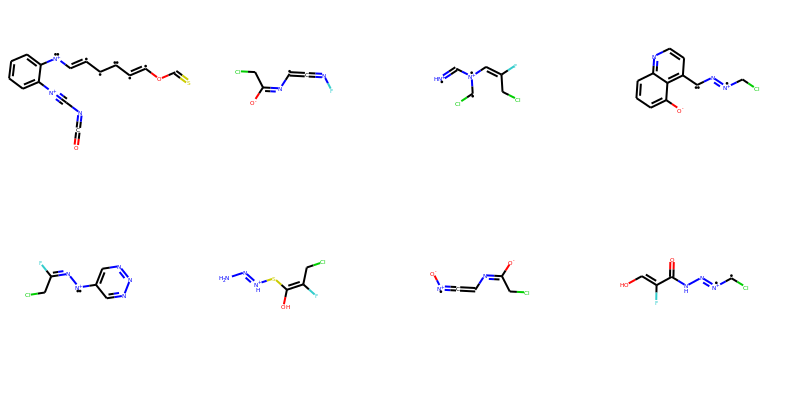}}
\end{subfigure}
\caption{\textbf{Visual comparison of exemplar inhibitors for assay $1,706$.} (a) from the dataset, (b) discovered by MCTS with and (c) without energy regularization. The inhibitors discovered with energy regularization visually resemble molecules found within the dataset. In contrast, those discovered without energy regularization are often simple and linear. }  \label{fig:mols}
 \end{center}
 \end{figure}

Figure~\ref{fig:mols} shows eight known inhibitors for assay $1,706$ in comparison to eight inhibitors discovered with and without energy regularization. This result clearly indicates the importance of the energy model in the reward function. The inhibitors discovered with DEEN enabled closely resemble molecules found within the dataset. In contrast, those inhibitors discovered without energy regularization ($\beta=0$) are often unusually simple and linear.

\begin{figure}
    \centering
    \includegraphics[width=0.8\textwidth]{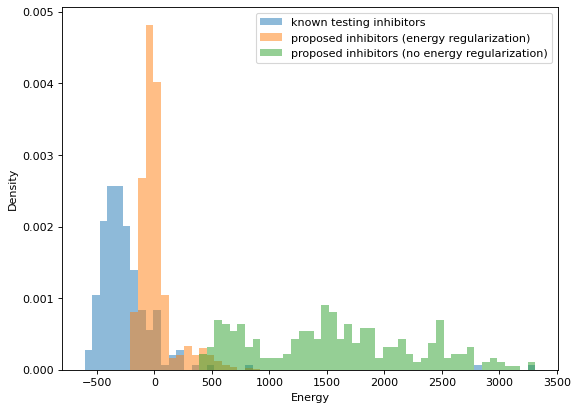}\\
    \caption{\textbf{Energies of the molecules discovered} with and without energy regularization for assay $1,706$, compare to those of known (testing) inhibitors.}
    \label{fig:mol_en}
\end{figure}

To highlight this point, Figure \ref{fig:mol_en} compares the energies (as computed by the energy model) of various discovered inhibitors with and without energy regularization, and known (testing) inhibitors. Many of the proposed inhibitors generated without energy regularization are far from the 
learned manifold of the known inhibitors, while the distribution of inhibitors discovered with regularization are centered within the range of known-inhibitor energy values. In fact, they are much more concentrated than the known inhibitors, suggesting that perhaps the best explored inhibitors lie within a similar subspace of the overall combinatorial space of molecules.

\begin{figure}
    \centering
    \includegraphics[width=0.8\textwidth]{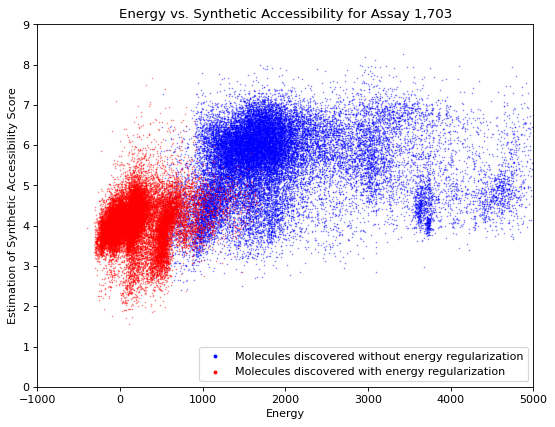}\\
    \caption{\textbf{Energy vs. estimation of synthetic accessibility} \cite{ertl2009estimation} with (red) and without (blue) energy regularization for assay $1,706$.}
    \label{fig:sa_vs_energy}
\end{figure}

Figure~\ref{fig:sa_vs_energy} demonstrates that the energy model has also captured the general `synthesizability' property of the original dataset. In this figure, the estimation of synthetic accessibility score~\cite{ertl2009estimation} of each discovered potential inhibitor is plotted against energy for assay $1,706$, both with (red) and without (blue) energy regularization. Molecules discovered with regularization tend to have substantially lower estimation of synthetic accessibility scores, meaning that it is more likely that they are practically synthesizable.

\section{Conclusion \label{sec:conclude}}
With the goal of designing novel inhibitors for SARS-CoV-1 and SARS-CoV-2, \framework brings together GNNs for discriminating 3CL\textsuperscript{pro} and PL\textsuperscript{pro} inhibitors based on a publicly available database, DEEN for learning the energy function of molecules in the feature space of the trained GNN, and MCTS for searching within the space of inhibitors. MCTS was guided via a novel reward function whose form was dictated by both the GNN's learned discriminator and DEEN's learned energy function. The use of an energy function here can be informally thought of as a form of soft constraint satisfaction\textemdash the ``constraint'' is to be statistically close to the space of molecules in the dataset\textemdash where both $\beta$ and $\sigma$ control the ``softness''. Apart from hyperparameter search, which is inherently problem specific, the machinery developed here is quite general, and could be applied to any labeled dataset of molecules for drug discovery in particular and in biotechnology at large.

\bibliographystyle{plainnat}
\bibliography{bibliography}

\begin{thebibliography}{56}
\providecommand{\natexlab}[1]{#1}
\providecommand{\url}[1]{\texttt{#1}}
\expandafter\ifx\csname urlstyle\endcsname\relax
  \providecommand{\doi}[1]{doi: #1}\else
  \providecommand{\doi}{doi: \begingroup \urlstyle{rm}\Url}\fi

\bibitem[Auer et~al.(2002)Auer, Cesa{-}Bianchi, and Fischer]{auer2002finite}
Peter Auer, Nicol{\`{o}} Cesa{-}Bianchi, and Paul Fischer.
\newblock Finite-time analysis of the multiarmed bandit problem.
\newblock \emph{Machine learning}, 47\penalty0 (2-3):\penalty0 235--256, 2002.

\bibitem[Battaglia et~al.(2018)Battaglia, Hamrick, Bapst, Sanchez-Gonzalez,
  Zambaldi, Malinowski, Tacchetti, Raposo, Santoro, Faulkner,
  et~al.]{battaglia2018relational}
Peter~W Battaglia, Jessica~B Hamrick, Victor Bapst, Alvaro Sanchez-Gonzalez,
  Vinicius Zambaldi, Mateusz Malinowski, Andrea Tacchetti, David Raposo, Adam
  Santoro, Ryan Faulkner, et~al.
\newblock Relational inductive biases, deep learning, and graph networks.
\newblock \emph{arXiv preprint arXiv:1806.01261}, 2018.

\bibitem[Beck et~al.(2020)Beck, Shin, Choi, Park, and Kang]{beck2020predicting}
Bo~Ram Beck, Bonggun Shin, Yoonjung Choi, Sungsoo Park, and Keunsoo Kang.
\newblock Predicting commercially available antiviral drugs that may act on the
  novel coronavirus (sars-cov-2) through a drug-target interaction deep
  learning model.
\newblock \emph{Computational and structural biotechnology journal},
  18:\penalty0 784 -- 790, 2020.

\bibitem[Breiman(1996)]{breiman1996bagging}
Leo Breiman.
\newblock Bagging predictors.
\newblock \emph{Machine learning}, 24\penalty0 (2):\penalty0 123--140, 1996.

\bibitem[{Bronstein} et~al.(2017){Bronstein}, {Bruna}, {LeCun}, {Szlam}, and
  {Vandergheynst}]{bronstein2017gdlsurvey}
M.~M. {Bronstein}, J.~{Bruna}, Y.~{LeCun}, A.~{Szlam}, and P.~{Vandergheynst}.
\newblock Geometric deep learning: Going beyond euclidean data.
\newblock \emph{IEEE Signal Processing Magazine}, 34\penalty0 (4):\penalty0
  18--42, 2017.

\bibitem[Browne et~al.(2012)Browne, Powley, Whitehouse, Lucas, Cowling,
  Rohlfshagen, Tavener, Perez, Samothrakis, and Colton]{browne2012survey}
Cameron~B Browne, Edward Powley, Daniel Whitehouse, Simon~M Lucas, Peter~I
  Cowling, Philipp Rohlfshagen, Stephen Tavener, Diego Perez, Spyridon
  Samothrakis, and Simon Colton.
\newblock A survey of monte carlo tree search methods.
\newblock \emph{IEEE Trans. Computational Intelligence and AI in games},
  4\penalty0 (1):\penalty0 1--43, 2012.

\bibitem[Butt et~al.(2020)Butt, Gill, Chun, and Babu]{butt2020deep}
Charmaine Butt, Jagpal Gill, David Chun, and Benson~A Babu.
\newblock Deep learning system to screen coronavirus disease 2019 pneumonia.
\newblock \emph{Applied Intelligence}, pages 1--7, 2020.

\bibitem[Cangea et~al.(2018)Cangea, Veli{\v{c}}kovi{\'c}, Jovanovi{\'c}, Kipf,
  and Li{\`o}]{cangea2018towards}
C{\u{a}}t{\u{a}}lina Cangea, Petar Veli{\v{c}}kovi{\'c}, Nikola Jovanovi{\'c},
  Thomas Kipf, and Pietro Li{\`o}.
\newblock Towards sparse hierarchical graph classifiers.
\newblock \emph{arXiv preprint arXiv:1811.01287}, 2018.

\bibitem[Cofala et~al.(2020)Cofala, Elend, Mirbach, Prellberg, Teusch, and
  Kramer]{cofala2020evolutionary}
Tim Cofala, Lars Elend, Philip Mirbach, Jonas Prellberg, Thomas Teusch, and
  Oliver Kramer.
\newblock Evolutionary multi-objective design of sars-cov-2 protease inhibitor
  candidates.
\newblock In \emph{Parallel Problem Solving from Nature -- PPSN XVI}, pages
  357--371. Springer, 2020.

\bibitem[Diehl(2019)]{diehl2019edge}
Frederik Diehl.
\newblock Edge contraction pooling for graph neural networks.
\newblock \emph{arXiv preprint arXiv:1905.10990}, 2019.

\bibitem[Duvenaud et~al.(2015)Duvenaud, Maclaurin, Iparraguirre, Bombarell,
  Hirzel, Aspuru-Guzik, and Adams]{DuvenaudMAGHAA15}
David~K Duvenaud, Dougal Maclaurin, Jorge Iparraguirre, Rafael Bombarell,
  Timothy Hirzel, Alan Aspuru-Guzik, and Ryan~P Adams.
\newblock Convolutional networks on graphs for learning molecular fingerprints.
\newblock In \emph{Advances in Neural Information Processing Systems},
  volume~28, pages 2224--2232, 2015.

\bibitem[Elfwing et~al.(2017)Elfwing, Uchibe, and Doya]{elfwing2017sigmoid}
Stefan Elfwing, Eiji Uchibe, and Kenji Doya.
\newblock Sigmoid-weighted linear units for neural network function
  approximation in reinforcement learning.
\newblock \emph{arXiv preprint arXiv:1702.03118}, 2017.

\bibitem[Ertl and Schuffenhauer(2009)]{ertl2009estimation}
Peter Ertl and Ansgar Schuffenhauer.
\newblock Estimation of synthetic accessibility score of drug-like molecules
  based on molecular complexity and fragment contributions.
\newblock \emph{Journal of cheminformatics}, 1\penalty0 (1):\penalty0 8, 2009.

\bibitem[Fischer et~al.(2020)Fischer, Sellner, Neranjan, Smieško, and
  Lill]{fischer2020potential}
André Fischer, Manuel Sellner, Santhosh Neranjan, Martin Smieško, and
  Markus~A. Lill.
\newblock Potential inhibitors for novel coronavirus protease identified by
  virtual screening of 606 million compounds.
\newblock \emph{International Journal of Molecular Sciences}, 21\penalty0 (10),
  2020.

\bibitem[Gawehn et~al.(2016)Gawehn, Hiss, and Schneider]{gawehn2016deep}
Erik Gawehn, Jan~A Hiss, and Gisbert Schneider.
\newblock Deep learning in drug discovery.
\newblock \emph{Molecular informatics}, 35\penalty0 (1):\penalty0 3--14, 2016.

\bibitem[Gilmer et~al.(2017)Gilmer, Schoenholz, Riley, Vinyals, and
  Dahl]{gilmer2017neural}
Justin Gilmer, Samuel~S Schoenholz, Patrick~F Riley, Oriol Vinyals, and
  George~E Dahl.
\newblock Neural message passing for quantum chemistry.
\newblock \emph{arXiv preprint arXiv:1704.01212}, 2017.

\bibitem[Goh et~al.(2017)Goh, Hodas, and Vishnu]{goh2017deep}
Garrett~B Goh, Nathan~O Hodas, and Abhinav Vishnu.
\newblock Deep learning for computational chemistry.
\newblock \emph{Journal of computational chemistry}, 38\penalty0 (16):\penalty0
  1291--1307, 2017.

\bibitem[Hamilton et~al.(2017)Hamilton, Ying, and
  Leskovec]{hamilton2017reprlearninggraphs}
William~L Hamilton, Rex Ying, and Jure Leskovec.
\newblock Representation learning on graphs: Methods and applications.
\newblock \emph{arXiv preprint arXiv:1709.05584}, 2017.

\bibitem[Hofmarcher et~al.(2020)Hofmarcher, Mayr, Rumetshofer, Ruch, Renz,
  Schimunek, Seidl, Vall, Widrich, Hochreiter, et~al.]{hofmarcher2020large}
Markus Hofmarcher, Andreas Mayr, Elisabeth Rumetshofer, Peter Ruch, Philipp
  Renz, Johannes Schimunek, Philipp Seidl, Andreu Vall, Michael Widrich, Sepp
  Hochreiter, et~al.
\newblock Large-scale ligand-based virtual screening for sars-cov-2 inhibitors
  using deep neural networks.
\newblock \emph{Available at SSRN 3561442}, 2020.

\bibitem[Hy et~al.(2018)Hy, Trivedi, Pan, Anderson, and
  Kondor]{hy2018predicting}
Truong~Son Hy, Shubhendu Trivedi, Horace Pan, Brandon~M Anderson, and Risi
  Kondor.
\newblock Predicting molecular properties with covariant compositional
  networks.
\newblock \emph{The Journal of chemical physics}, 148\penalty0 (24):\penalty0
  241745, 2018.

\bibitem[Hyv{\"a}rinen(2005)]{hyvarinen2005estimation}
Aapo Hyv{\"a}rinen.
\newblock Estimation of non-normalized statistical models by score matching.
\newblock \emph{Journal of Machine Learning Research}, 6\penalty0
  (Apr):\penalty0 695--709, 2005.

\bibitem[(JEDI)()]{JEDI}
Joint European Disruptive~Initiative (JEDI).
\newblock Jedi grand challence stage 1: Ultimate list of lead compounds by
  screening a billion molecules against covid-19.
\newblock \url{https://www.covid19.jedi.group/step-1}.

\bibitem[Jensen(2019)]{jensen2019graph}
Jan~H Jensen.
\newblock A graph-based genetic algorithm and generative model/monte carlo tree
  search for the exploration of chemical space.
\newblock \emph{Chemical science}, 10\penalty0 (12):\penalty0 3567--3572, 2019.

\bibitem[Kim et~al.(2019)Kim, Chen, Cheng, Gindulyte, He, He, Li, Shoemaker,
  Thiessen, Yu, et~al.]{kim2019pubchem}
Sunghwan Kim, Jie Chen, Tiejun Cheng, Asta Gindulyte, Jia He, Siqian He,
  Qingliang Li, Benjamin~A Shoemaker, Paul~A Thiessen, Bo~Yu, et~al.
\newblock Pubchem 2019 update: improved access to chemical data.
\newblock \emph{Nucleic acids research}, 47\penalty0 (D1):\penalty0
  D1102--D1109, 2019.

\bibitem[Kingma and Ba(2014)]{kingma2014adam}
Diederik~P Kingma and Jimmy Ba.
\newblock Adam: A method for stochastic optimization.
\newblock \emph{arXiv preprint arXiv:1412.6980}, 2014.

\bibitem[Kingma and Welling(2013)]{kingma2013auto}
Diederik~P Kingma and Max Welling.
\newblock Auto-encoding variational {B}ayes.
\newblock \emph{arXiv preprint arXiv:1312.6114}, 2013.

\bibitem[Kipf et~al.(2018)Kipf, Fetaya, Wang, Welling, and
  Zemel]{kipf2018neural}
Thomas Kipf, Ethan Fetaya, Kuan-Chieh Wang, Max Welling, and Richard Zemel.
\newblock Neural relational inference for interacting systems.
\newblock \emph{arXiv preprint arXiv:1802.04687}, 2018.

\bibitem[Kipf and Welling(2017)]{kipf2017semi}
Thomas~N. Kipf and Max Welling.
\newblock Semi-supervised classification with graph convolutional networks.
\newblock In \emph{International Conference on Learning Representations}, 2017.

\bibitem[Kocsis and Szepesv{\'a}ri(2006)]{kocsis2006bandit}
Levente Kocsis and Csaba Szepesv{\'a}ri.
\newblock Bandit based monte-carlo planning.
\newblock In \emph{European conference on machine learning}, pages 282--293.
  Springer, 2006.

\bibitem[Kraev(2018)]{kraev2018grammars}
Egor Kraev.
\newblock Grammars and reinforcement learning for molecule optimization.
\newblock \emph{arXiv preprint arXiv:1811.11222}, 2018.

\bibitem[Landrum()]{rdkit}
Greg Landrum.
\newblock Rdkit: Open-source cheminformatics.
\newblock \url{http://www.rdkit.org}.

\bibitem[{Latif} et~al.(2020){Latif}, {Usman}, {Manzoor}, {Iqbal}, {Qadir},
  {Tyson}, {Castro}, {Razi}, {Boulos}, {Weller}, and
  {Crowcroft}]{latif2020leveraging}
S.~{Latif}, M.~{Usman}, S.~{Manzoor}, W.~{Iqbal}, J.~{Qadir}, G.~{Tyson},
  I.~{Castro}, A.~{Razi}, M.~N.~K. {Boulos}, A.~{Weller}, and J.~{Crowcroft}.
\newblock Leveraging data science to combat covid-19: A comprehensive review.
\newblock \emph{IEEE Trans. Artificial Intelligence}, 1\penalty0 (1):\penalty0
  85--103, 2020.

\bibitem[Lee et~al.(2019)Lee, Lee, and Kang]{lee2019self}
Junhyun Lee, Inyeop Lee, and Jaewoo Kang.
\newblock Self-attention graph pooling.
\newblock \emph{arXiv preprint arXiv:1904.08082}, 2019.

\bibitem[Macchiagodena et~al.(2020)Macchiagodena, Pagliai, and
  Procacci]{macchiagodena2020inhibition}
Marina Macchiagodena, Marco Pagliai, and Piero Procacci.
\newblock Inhibition of the main protease 3cl-pro of the coronavirus disease 19
  via structure-based ligand design and molecular modeling.
\newblock \emph{arXiv preprint arXiv:2002.09937}, 2020.

\bibitem[Mayr et~al.(2018)Mayr, Klambauer, Unterthiner, Steijaert, Wegner,
  Ceulemans, Clevert, and Hochreiter]{mayr2018large}
Andreas Mayr, G{\"u}nter Klambauer, Thomas Unterthiner, Marvin Steijaert,
  J{\"o}rg~K Wegner, Hugo Ceulemans, Djork-Arn{\'e} Clevert, and Sepp
  Hochreiter.
\newblock Large-scale comparison of machine learning methods for drug target
  prediction on chembl.
\newblock \emph{Chemical science}, 9\penalty0 (24):\penalty0 5441--5451, 2018.

\bibitem[Miyasawa(1961)]{miyasawa1961empirical}
Koichi Miyasawa.
\newblock An empirical {B}ayes estimator of the mean of a normal population.
\newblock \emph{Bulletin of the International Statistical Institute},
  38\penalty0 (4):\penalty0 181--188, 1961.

\bibitem[Perozzi et~al.(2014)Perozzi, Al-Rfou, and Skiena]{perozzi2014dw}
Bryan Perozzi, Rami Al-Rfou, and Steven Skiena.
\newblock Deepwalk: Online learning of social representations.
\newblock In \emph{Proc. 20th ACM SIGKDD International Conference on Knowledge
  Discovery and Data Mining}, page 701–710. Association for Computing
  Machinery, 2014.

\bibitem[Polishchuk et~al.(2013)Polishchuk, Madzhidov, and
  Varnek]{polishchuk2013estimation}
Pavel~G Polishchuk, Timur~I Madzhidov, and Alexandre Varnek.
\newblock Estimation of the size of drug-like chemical space based on gdb-17
  data.
\newblock \emph{Journal of computer-aided molecular design}, 27\penalty0
  (8):\penalty0 675--679, 2013.

\bibitem[Ramachandran et~al.(2017)Ramachandran, Zoph, and
  Le]{ramachandran2017swish}
Prajit Ramachandran, Barret Zoph, and Quoc~V Le.
\newblock Swish: a self-gated activation function.
\newblock \emph{arXiv preprint arXiv:1710.05941}, 7, 2017.

\bibitem[Robbins(1956)]{robbins1956empirical}
Herbert Robbins.
\newblock An empirical {B}ayes approach to statistics.
\newblock In \emph{Proc. Third Berkeley Symp.}, volume~1, pages 157--163, 1956.

\bibitem[Saremi and Hyv{\"a}rinen(2019)]{saremi2019neural}
Saeed Saremi and Aapo Hyv{\"a}rinen.
\newblock Neural empirical {B}ayes.
\newblock \emph{Journal of Machine Learning Research}, 20:\penalty0 1--23,
  2019.

\bibitem[Saremi et~al.(2018)Saremi, Mehrjou, Sch{\"o}lkopf, and
  Hyv{\"a}rinen]{saremi2018deep}
Saeed Saremi, Arash Mehrjou, Bernhard Sch{\"o}lkopf, and Aapo Hyv{\"a}rinen.
\newblock Deep energy estimator networks.
\newblock \emph{arXiv preprint arXiv:1805.08306}, 2018.

\bibitem[Scarselli et~al.(2009)Scarselli, Gori, Tsoi, Hagenbuchner, and
  Monfardini]{Scarselli2009TheGN}
Franco Scarselli, Marco Gori, Ah~Chung Tsoi, Markus Hagenbuchner, and Gabriele
  Monfardini.
\newblock The graph neural network model.
\newblock \emph{IEEE Trans. Neural Networks}, 20:\penalty0 61--80, 2009.

\bibitem[St.~John et~al.(2019)St.~John, Phillips, Kemper, Wilson, Guan,
  Crowley, Nimlos, and Larsen]{st2019message}
Peter~C St.~John, Caleb Phillips, Travis~W Kemper, A~Nolan Wilson, Yanfei Guan,
  Michael~F Crowley, Mark~R Nimlos, and Ross~E Larsen.
\newblock Message-passing neural networks for high-throughput polymer
  screening.
\newblock \emph{The Journal of chemical physics}, 150\penalty0 (23):\penalty0
  234111, 2019.

\bibitem[St{\aa}hl et~al.(2019)St{\aa}hl, Falkman, Karlsson, Mathiason, and
  Bostrom]{staahl2019deep}
Niclas St{\aa}hl, Goran Falkman, Alexander Karlsson, Gunnar Mathiason, and
  Jonas Bostrom.
\newblock Deep reinforcement learning for multiparameter optimization in de
  novo drug design.
\newblock \emph{Journal of Chemical Information and Modeling}, 59\penalty0
  (7):\penalty0 3166--3176, 2019.

\bibitem[Tang et~al.(2020)Tang, He, Liu, Fang, Wu, and Xu]{tang2020ai}
Bowen Tang, Fengming He, Dongpeng Liu, Meijuan Fang, Zhen Wu, and Dong Xu.
\newblock Ai-aided design of novel targeted covalent inhibitors against
  sars-cov-2.
\newblock \emph{bioRxiv}, 2020.

\bibitem[Ton et~al.(2020)Ton, Gentile, Hsing, Ban, and Cherkasov]{ton2020rapid}
Anh-Tien Ton, Francesco Gentile, Michael Hsing, Fuqiang Ban, and Artem
  Cherkasov.
\newblock Rapid identification of potential inhibitors of sars-cov-2 main
  protease by deep docking of 1.3 billion compounds.
\newblock \emph{Molecular informatics}, 2020.

\bibitem[Wainwright and Jordan(2008)]{wainwright2008graphical}
Martin~J Wainwright and Michael~I Jordan.
\newblock Graphical models, exponential families, and variational inference.
\newblock \emph{Foundations and Trends in Machine Learning}, 1\penalty0
  (1--2):\penalty0 1--305, 2008.

\bibitem[Wang et~al.(2019)Wang, Sun, Liu, Sarma, Bronstein, and
  Solomon]{wang2018dyngraph}
Yue Wang, Yongbin Sun, Ziwei Liu, Sanjay~E. Sarma, Michael~M. Bronstein, and
  Justin~M. Solomon.
\newblock Dynamic graph {CNN} for learning on point clouds.
\newblock \emph{ACM Trans. Graphics}, 38\penalty0 (5), 2019.

\bibitem[Weininger(1988)]{weininger1988smiles}
David Weininger.
\newblock Smiles, a chemical language and information system. 1. introduction
  to methodology and encoding rules.
\newblock \emph{Journal of chemical information and computer sciences},
  28\penalty0 (1):\penalty0 31--36, 1988.

\bibitem[Xiong et~al.(2020)Xiong, Wang, Liu, Zhong, Wan, Li, Li, Luo, Chen,
  Jiang, et~al.]{xiong2019pushing}
Zhaoping Xiong, Dingyan Wang, Xiaohong Liu, Feisheng Zhong, Xiaozhe Wan, Xutong
  Li, Zhaojun Li, Xiaomin Luo, Kaixian Chen, Hualiang Jiang, et~al.
\newblock Pushing the boundaries of molecular representation for drug discovery
  with the graph attention mechanism.
\newblock \emph{Journal of Medicinal Chemistry}, 63\penalty0 (16):\penalty0
  8749--8760, 2020.

\bibitem[Yang et~al.(2019)Yang, Swanson, Jin, Coley, Eiden, Gao, Guzman-Perez,
  Hopper, Kelley, Mathea, et~al.]{yang2019analyzing}
Kevin Yang, Kyle Swanson, Wengong Jin, Connor Coley, Philipp Eiden, Hua Gao,
  Angel Guzman-Perez, Timothy Hopper, Brian Kelley, Miriam Mathea, et~al.
\newblock Analyzing learned molecular representations for property prediction.
\newblock \emph{Journal of chemical information and modeling}, 59\penalty0
  (8):\penalty0 3370--3388, 2019.

\bibitem[Zhang et~al.(2020)Zhang, Saravanan, Yang, Hossain, Li, Ren, Pan, and
  Wei]{zhang2020deep}
Haiping Zhang, Konda~Mani Saravanan, Yang Yang, Md~Tofazzal Hossain, Junxin Li,
  Xiaohu Ren, Yi~Pan, and Yanjie Wei.
\newblock Deep learning based drug screening for novel coronavirus 2019-ncov.
\newblock \emph{Interdisciplinary Sciences: Computational Life Sciences},
  12:\penalty0 368--376, 2020.

\bibitem[Zhang et~al.(2019)Zhang, Bu, Ester, Zhang, Yao, Yu, and
  Wang]{zhang2019hierarchical}
Zhen Zhang, Jiajun Bu, Martin Ester, Jianfeng Zhang, Chengwei Yao, Zhi Yu, and
  Can Wang.
\newblock Hierarchical graph pooling with structure learning.
\newblock \emph{arXiv preprint arXiv:1911.05954}, 2019.

\bibitem[Zhavoronkov et~al.(2020)Zhavoronkov, Zagribelnyy, Zhebrak, Aladinskiy,
  Terentiev, Vanhaelen, Bezrukov, Polykovskiy, Shayakhmetov, Filimonov, Bishop,
  McCloskey, Leija, Bright, Funakawa, Lin, Huang, Liao, Aliper, and
  Ivanenkov]{zhavoronkov2020potential}
Alex Zhavoronkov, Bogdan Zagribelnyy, Alexander Zhebrak, Vladimir Aladinskiy,
  Victor Terentiev, Quentin Vanhaelen, Dmitry Bezrukov, Daniil Polykovskiy, Rim
  Shayakhmetov, Andrey Filimonov, Michael Bishop, Steve McCloskey, Edgardo
  Leija, Deborah Bright, Keita Funakawa, Yen-Chu Lin, Shih-Hsien Huang,
  Hsuan-Jen Liao, Alex Aliper, and Yan Ivanenkov.
\newblock Potential non-covalent sars-cov-2 3c-like protease inhibitors
  designed using generative deep learning approaches and reviewed by human
  medicinal chemist in virtual reality.
\newblock 2020.

\bibitem[Zhou et~al.(2018)Zhou, Cui, Zhang, Yang, Liu, Wang, Li, and
  Sun]{zhou2018graph}
Jie Zhou, Ganqu Cui, Zhengyan Zhang, Cheng Yang, Zhiyuan Liu, Lifeng Wang,
  Changcheng Li, and Maosong Sun.
\newblock Graph neural networks: A review of methods and applications.
\newblock \emph{arXiv preprint arXiv:1812.08434}, 2018.

\end{thebibliography}

\pagebreak
\appendix
\section{Parameters for GNN Variants in compared in Table~\ref{tab:gnn_compare}}.

All GNN variants were trained using the Adam optimizer and the loss function described in Section \ref{sec:loss}. Each ensemble of networks is trained using the same five folds of training data, with each member of the ensemble using a different fold as a validation set for early stopping. In each case the final model chosen is that which achieved the lowest loss on its validation set, and early stopping is determined when the validation loss has not decreased for $20$ epochs. For each network variant, the hyper-parameters were chosen by hand-tuning the parameter settings that were either described in the respective publications or found in authors' publicly available code. 

\subsection{MPNN\cite{gilmer2017neural} Parameters}


\begin{itemize}
    \item Node dimensionality $=200$
    \item Edge dimensionality $=50$
    \item Layers $T=3$
    \item Number of Set2Set applications $=3$
    \item Batch size = 20
    \item Learning rate = $10^{-2.9}$
\end{itemize}

Every $50000$ batches, the learning rate is halved. 

\subsection{D-MPNN\cite{yang2019analyzing} Parameters}


\begin{itemize}
    \item Hidden dimensionality $=300$
    \item Number of Directed Message Passing layers $T=6$
    \item Output MLP layers $F=6$
    \item Batch size = 50
    \item Learning rate = $10^{-4}$
\end{itemize}

 Every $50000$ batches, the learning rate is halved. 

\subsection{Attentive FP\cite{xiong2019pushing} Parameters}

\begin{itemize}
    \item Radius $=3$
    \item $T = 2$
    \item Fingerprint dimensionality $=150$
    \item Dropout $p = 0.1$
    \item Batch size = $100$
    \item Learning rate = $10^{-3.5}$ 
    \item Weight decay = $10^{-2.9}$
\end{itemize}

\subsection{Graph Network\cite{battaglia2018relational} Parameters}


\begin{itemize}
    \item Node, edge and global dimensionality $=96$
    \item Number of message passing layers $T=3$
    \item Output MLP layers $F=2$
    \item Dropout on output layers $p=0.25$ and $p=0.5$, respectively
    \item Batch size = $128$,
    \item Learning rate = $2\times 10^{-5}$ 
    \item Weight decay = $10^{-4}$
\end{itemize}

\end{document}